\documentclass[letterpaper]{article} % DO NOT CHANGE THIS
\usepackage{aaai24}  % DO NOT CHANGE THIS
\usepackage{times}  % DO NOT CHANGE THIS
\usepackage{helvet}  % DO NOT CHANGE THIS
\usepackage{courier}  % DO NOT CHANGE THIS
\usepackage[hyphens]{url}  % DO NOT CHANGE THIS
\usepackage{graphicx} % DO NOT CHANGE THIS
\usepackage{natbib}  % DO NOT CHANGE THIS AND DO NOT ADD ANY OPTIONS TO IT
\usepackage{caption} % DO NOT CHANGE THIS AND DO NOT ADD ANY OPTIONS TO IT
\usepackage{algorithm}
\usepackage{algorithmic}
\usepackage[namelimits]{amsmath} 
\usepackage{amssymb}
\usepackage{amsfonts}
\usepackage{mathrsfs}
\usepackage{amsmath}
\usepackage{multirow}
\usepackage{booktabs}
\usepackage{subcaption}
\usepackage{bbding}
\usepackage{newfloat}
\usepackage{listings}
\DeclareCaptionStyle{ruled}{labelfont=normalfont,labelsep=colon,strut=off} % DO NOT CHANGE THIS
\floatstyle{ruled}
\newfloat{listing}{tb}{lst}{}
\floatname{listing}{Listing}
\title{Say Anything with Any Style}
\author{
	Shuai Tan\textsuperscript{\rm 1},
	Bin Ji\textsuperscript{\rm 1},
	Yu Ding\textsuperscript{\rm 2},
	Ye Pan\textsuperscript{\rm 1}\thanks{Corresponding author.}
}
\affiliations{
	\textsuperscript{\rm 1} Shanghai Jiao Tong University \\
	\textsuperscript{\rm 2} Virtual Human Group, Netease Fuxi AI Lab \\

	\{tanshuai0219, bin.ji, whitneypanye\}@sjtu.edu.cn, dingyu01@corp.netease.com 
}

\usepackage{bibentry}
\begin{document}
	
	\maketitle
	
	\begin{abstract}
		Generating stylized talking head with diverse head motions is crucial for achieving natural-looking videos but still remains challenging. Previous works either adopt a regressive method to capture the speaking style, resulting in a coarse style that is averaged across all training data, or employ a universal network to synthesize videos with different styles which causes suboptimal performance. To address these, we propose a novel dynamic-weight method, namely \textbf{S}ay \textbf{A}nything with \textbf{A}ny \textbf{S}tyle (SAAS), which queries the discrete style representation via a generative model with a learned style codebook. Specifically, we develop a multi-task VQ-VAE that incorporates three closely related tasks to learn a style codebook as a prior for style extraction. This discrete prior, along with the generative model, enhances the precision and robustness when extracting the speaking styles of the given style clips. By utilizing the extracted style, a residual architecture comprising a canonical branch and style-specific branch is employed to predict the mouth shapes conditioned on any driving audio while transferring the speaking style from the source to any desired one. To adapt to different speaking styles, we steer clear of employing a universal network by exploring an elaborate HyperStyle to produce the style-specific weights offset for the style branch. Furthermore, we construct a pose generator and a pose codebook to store the quantized pose representation, allowing us to sample diverse head motions aligned with the audio and the extracted style. Experiments demonstrate that our approach surpasses state-of-the-art methods in terms of both lip-synchronization and stylized expression. Besides, we extend our SAAS to video-driven style editing field and achieve satisfactory performance.
	\end{abstract}
	
	\section{Introduction}
	
	% 1. Talking face 有很多应用，目前都实现的是嘴型，但是style很少有人考虑
	Talking face generation has gained significant popularity due to its wide range of applications in virtual reality, film production, and games~\cite{pataranutaporn2021ai}. While significant efforts have been dedicated to generating synchronized lip motions~\cite{vougioukas2020realistic, tian2019audio2face} and rhythmic head movements~\cite{chen2020talking, zhang20213d}, the stylized expression, which plays a crucial role in conveying communicative information~\cite{PaulEkman2005WhatTF}, is often overlooked in most existing approaches.
	
\begin{figure}[t]
		\centering
		\includegraphics[width=1\linewidth]{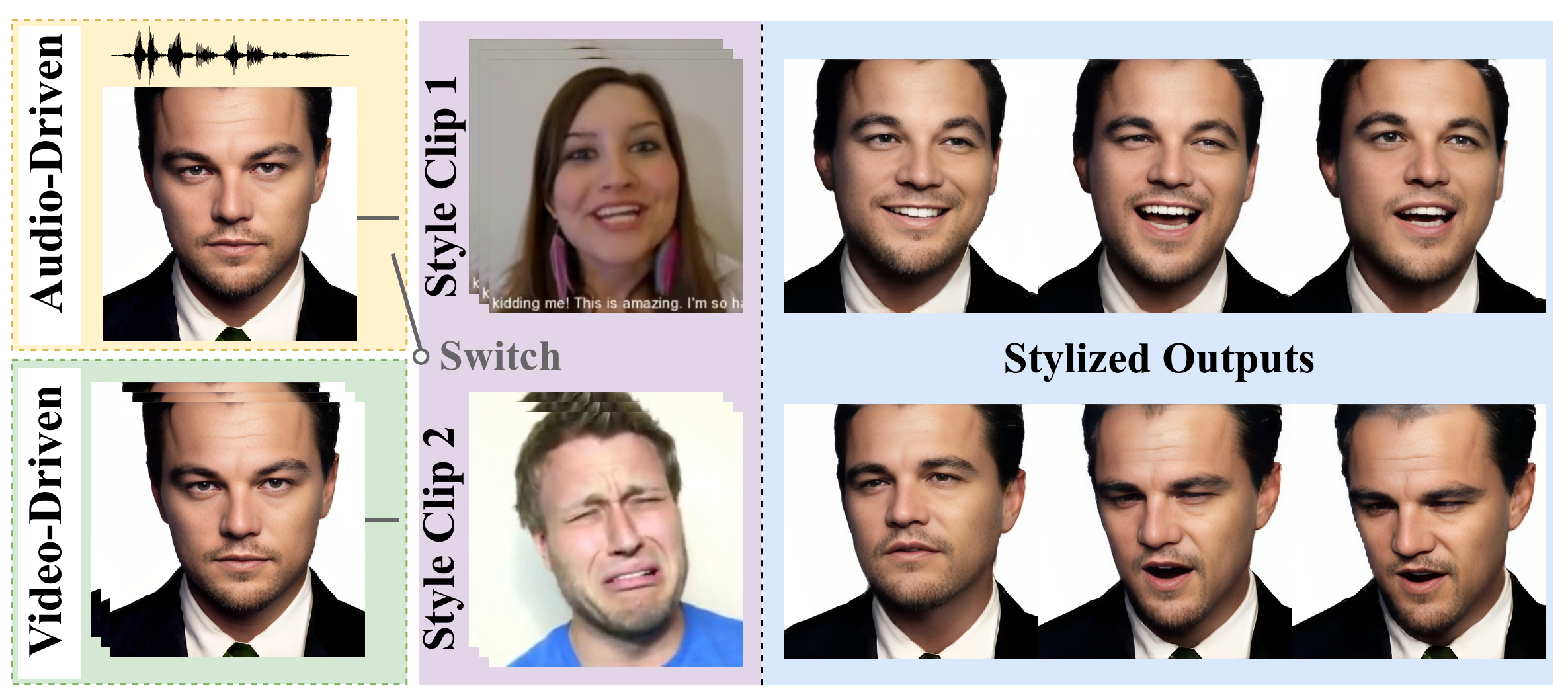}
		\caption{Example animations generated by our SAAS. Given a source image and a style reference clip, SAAS generates stylized talking faces driven by audio. The lip motions are synchronized with the audio, while the speaking styles are controlled by the style clips. We also support video-driven style editing by inputting a source video.}
		\label{fig:teaser}
	\end{figure}

	Generally speaking, individuals exhibit diverse speaking styles accompanied by corresponding variations in head poses when uttering the same sentence. As depicted in Figure~\ref{fig:teaser}, \textit{Leo} speaks with happy and sad styles, where the happier style often involves more lively head movements, whereas the sadder style tends to display a contrasting pattern. Previous methods~\cite{ji2022eamm,ma2023styletalk} commonly treat the style extraction as a regression task, inadvertently encouraging averaged representations and limiting the diversity of stylized expressions. Also, these methods use a static network to stylize motion for different styles which leads to suboptimal results. Moreover, they rarely consider the stylized head poses and instead rely on users to provide a pose reference.
	
	% 3. overview
	In this paper, we introduce a novel model called \textbf{S}ay \textbf{A}nything with \textbf{A}ny \textbf{S}tyle (\textbf{SAAS}). Our objective is to generate talking face videos with stylized expressions and head poses that resemble the provided style clip, while ensuring synchronization of lip motions with the audio. Additionally, we extend our method to video-driven style editing field, enabling the transfer of style from input videos to match a specified style clip. We leverage the 3DMM (3D Morphable Models)~\cite{blanz1999morphable} coefficients as the intermediate representation. Unlike previous regressive approach~\cite{ji2022eamm, ma2023styletalk}, we formulate style extraction as a query task using a learned style codebook, which expands a space for discrete style prior through a generative model, improving the accuracy and robustness to the unseen style extraction. To facilitate this, we propose a multi-task VQ-VAE to jointly learn the codebook and a style encoder. This differs from the original VQ-VAE~\cite{van2017neural} by incorporating three highly-related tasks instead of solely focusing on reconstruction. Concretely, the main style extraction task encodes discrete speaking styles from the style clip, while two auxiliary tasks, namely sequence reconstruction and style classification, are designed to enhance the style encoder and codebook to learn dynamic and explicit speaking styles. By utilizing the style codebook as a prior which encompasses various forms of style in a discrete format, we effectively condense an arbitrary style clip into a combination of the most pertinent codebook elements. Consequently, the robustness of subsequent modules in our framework is also significantly enhanced, since the extracted style is dragged closer to the seen style of training dataset.

	To perform stylized talking face, we devise a residual architecture~\cite{rebuffi2017learning} consisting of two branches: a canonical branch and a style-specific branch. The canonical branch is responsible for predicting lip motion and accommodating diverse styles into canonical, thereby facilitating the transfer between any two styles. On the other hand, the style-specific branch generates stylized displacements to the canonical representation. By innovatively treating the different style generation as domain adaptation, one straightforward idea is to design a specific branch for a new style~\cite{mason2018few}. However, it is non-trivial to achieve due to the immense number of possible styles that overload computational resources and the inability to handle unseen talking styles. To this end, we introduce a HyperStyle to modulate the weights of a single style-specific branch with the guidance of the extracted speaking style. In this fashion, we not only reduce the requirement for multiple branches but also enable generalizability for arbitrary stylization. Lastly, by combining representations from both branches, our model complementarily achieves stylized facial motion generation with lip-synchronization.
	
	% 6. pose
	As for stylized head motions, we create a pose codebook to store quantized pose representation. Since the extracted style indicates the impact of stylized expressions on head poses, we develop a cross-modal pose generator that maps from the speaking style and the driving audio to a distribution of pose quantization, from which diverse head poses can be sampled. Besides, the learned discrete latent codes in pose codebook remains within the realm of realistic head motion. This guarantees the stability and coherence of the generated motions in long-term predictions. Next, a Face Render with facial discriminators is adapted to generate stylized videos from predicted expression and head pose coefficients. Extensive experiments demonstrate the superiority of our method compared to state-of-the-arts (SOTAs).
	
	Our contributions are summarized as follows:
	\begin{itemize}
		\item We propose \textbf{S}ay \textbf{A}nything with \textbf{A}ny \textbf{S}tyle model (i.e., \textbf{SAAS}) to generate accurate lip motion synchronized with audio, and stylized expressions and head motions imitating any style clip. Besides, we extend our method to challenging video-driven style editing task.
		\item By taking advantage of discrete representation learning, we learn a style codebook by designing a multi-task VQ-VAE to extract a more explicit speaking style. Another pose codebook and pose generator are constructed to generate diverse stylized head motion sequences. 
		\item Our proposed HyperStyle effectively reduces the branch required for each new style, allowing a single style-specific branch to transfer the source to arbitrary style and further generate stylized videos.
	\end{itemize}

	\section{Related Work}
	\subsection{Audio-driven Talking Face Generation}
	
	As deep learning advances, the generated outcomes have become increasingly impressive. Early methods~\cite{alghamdi2022talking, zhou2019talking} primarily focus on achieving lip synchronization with the input audio. To enhance the naturalness and realism of the results, recent studies~\cite{zhang2023dinet, zhang2022sadtalker} take the head pose into account. Some works draw on the intermediate representation to bridge the gap between audio and video modality, such as landmark~\cite{chen2019hierarchical, zhou2020makelttalk} and dense motion field~\cite{wang2021audio2head, wang2022one}. Another popular framework~\cite{prajwal2020lip, park2022synctalkface, zhou2021pose} involves encoding and decoding for feature fusion and video reconstructing. However, these approaches rarely synthesize stylized results.
	
	To incorporate expressive facial expressions into talking face videos for more vivid performance, several approaches~\cite{SanjanaSinha2022EmotionControllableGT, ji2021audio, li2021write, tan2023emmn, pan2024expressive, pan2023emotional} introduce one-hot vectors representing common emotions as additional input to generate emotional talking face videos. However, relying solely on discrete emotion labels limits the amount of information available about the nuanced expression, ultimately reducing the diversity of the generated results~\cite{ji2022eamm}. More recently, alternative methods~\cite{liang2022expressive, ji2022eamm, ma2023styletalk} propose generating expressive talking heads by transferring expressions from an additional emotional source video to the target speaker. Nevertheless, these methods treat style extracting as a regression task and utilize a universal network to process different styles, which leads to suboptimal motions. In contrast, we quantize the latent style features stored in a specially designed codebook and design style-specific network, which facilitates generated videos with a more distinctive speaking style.
	
	\begin{figure*}[t]
		\centering
		\includegraphics[width=0.92\linewidth]{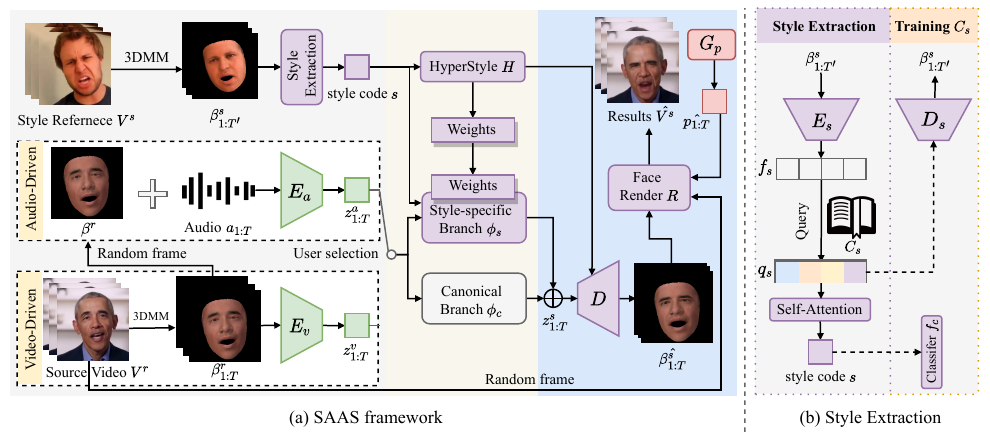}
		\caption{(a) The overview of SAAS. We first extract expression coefficients $\beta^s_{1:T'}$ from style reference video $V^s$ by 3DMM and extract the style code $s$. Audio Encoder $E_a$ encodes coefficient $\beta^r$ of source image and driving audio $a_{1:T}$ into $z^a_{1:T}$, which is fed into canonical branch $\phi_c$ and style-specific branch $\phi_s$. To generate stylized motion, $\phi_s$ accept the style-specific weights produced by HyperStyle $H$ and transfer $z^a_{1:T}$ into stylized $z^s_{1:T}$. Decoder $D$ reconstructs the coefficients $\hat{\beta^s_{1:T}}$ and Face Render $R$ synthesise the stylized video $\hat{V^s}$ along with the predicted head pose $\hat{p_{1:T}}$ by proposed Pose Generator $G_p$. (b) The pipeline of Style Extraction. The dotted arrow indicates the processes in $C_s$ training phase.}
		\label{fig:overview}
	\end{figure*}
	
	\subsection{Discrete Representation Learning}
	% CodeTalker 那部分的文章
	In recent times, discrete representation learning has yielded successful results in image restoration~\cite{jo2021practical} and generative task~\cite{dieleman2018challenge}. Among the methods for discrete representation learning, the VQ-VAE approach~\cite{van2017neural} has gained significant popularity for quantizing latent features into a learned codebook. \citet{ng2022learning} and \citet{xing2023codetalker} store the discrete prior of facial motion for more accurate movement. However, storing stylized discrete representation in codebook has yet to be attempted. Sparked by their approaches, we explore a multi-task VQ-VAE to extract the speaking style with the help of a learned codebook.

	\subsection{HyperNetwork}
	HyperNetwork~\cite{ha2016hypernetworks} leverages one auxiliary network, known as hypernetwork, to generate weights for a main network. By generating input-specific weights, HyperNetwork has demonstrated remarkable effectiveness in various domains~\cite{zamora2019adaptive, chen2020dynamic}. \citet{ye2022audio} propose a dynamic convolution kernel adjustment for a U-NET-like~\cite{ronneberger2015u} network based on the input audio, enabling the generation of talking videos frame-by-frame. However, directly generating weights from audio per frame poses challenges in capturing temporal relations among frames. Differently, our intuition is to predict offsets for the weights of the style-specific branch according to the extracted speaking style. In this fashion, we can modulate the branch to not only faithfully generate the corresponding stylized expressions but also retain the capability of the original structure to process temporal information effectively.

	\section{Proposed Method}
	% overview
	Our proposed framework, named \textbf{SAAS} (\textbf{S}ay \textbf{A}nything with \textbf{A}ny \textbf{S}tyle), aims to synthesize stylized talking head videos, whose identity, lip motion and stylized expression are consistent with different kinds of inputs: the reference image, audio speech and style video clip, respectively. Figure~\ref{fig:overview} illustrates the overview of our SAAS. The process begins with extracting a discrete speaking style representation $s$ from the stylized video $V^s$ via proposed Style Extraction module. Subsequently, A two-branch ($\phi_c$\&$\phi_s$) stylized module with HyperStyle $H$ is introduced to generate stylized expression coefficients $\hat{\beta^s_{1:T}}$ conditioned on the style $s$ and audio $a_{1:T}$. To enhance realism, we adopt a facial-enhanced Face Render $R$ to generate the final video $\hat{V^s}$ based on generated coefficients. Additionally, we extend our SAAS to transfer the speaking style of the input video $V^r$ to the extracted style $s$ while maintaining the lip motion unchanged.

	\subsection{Discrete Speaking Style Representation}
	% \label{style_encoder}
	In this paper, we employ 3D reconstruction model~\cite{deng2019accurate} to extract 3DMM coefficients from the input images and videos, in which the coefficients $\beta \in \mathbb{R}^{64}$ and $p \in \mathbb{R}^{6}$ describe the expression and head pose, respectively. By utilizing 3DMM, we not only exploit the 3D spatial information which is essential to capture facial motions, but also circumvent the effects of irrelevant appearance and illumination. For the sake of brevity, we omit the extraction process of 3DMM parameters in the rest of the writing.

	As presented in Figure~\ref{fig:overview} (b), we adopt the VQ-VAE framework to obtain the discretized latent space of speaking style stored in a style codebook, jointly with training an encoder and self-attention layer to embed 3D coefficients into style code. Note that there exist several significantly improved modifications. First, to capture dynamic speaking style, a transformer-based encoder $E_s$ is introduced to take the temporal information into account. Second, the goal of our VQ-VAE is to learn explicit speaking style representation instead of producing realistic facial motions. Therefore, we explore the multi-task learning consisting of three highly-related tasks: a main task for extracting style code, and two auxiliary tasks for reconstructing the input sequence and classifying its speaking style, respectively. The three related tasks are integrated via a shared encoder and codebook, following the multi-task learning strategy~\cite{tan2021incorporating, wen2023transformer}. Thanks to the modifications, we enhance the extraction of speaking style contained in the style clip, while eliminating the effect of the content information contained in the style clip~\cite{ji2022eamm} on the synchronization of the speech and audio. Specifically, the style coefficient sequence $\beta^s_{1:T'}$ are first embedded into a style feature $f_s = E_s(\beta^s_{1:T'}) \in \mathbb{R}^{\tau \times d_s} $, $\tau = \frac{T'}{\omega}$, where $T'$ and $\omega$ donate the length of style clip and temporal window, respectively. Then $f_s$ queries the style codebook $C_s \in \mathbb{R}^{N \times d_s}$ to retrieve its closest entry $q_s \in \mathbb{R}^{\tau \times d_s}$:
	
	\begin{equation}
		q_s=q(f_s):=\underset{{c_s}_k \in {C_s}}{\arg \min }\left\|{f_s}-{c_s}_k\right\|_2,
	\end{equation}
	which is fed into the self-attention~\cite{safari2020self} layer and decoder $D_s$ to extract the style code $s$ and reconstructing $\beta^s_{1:T'}$, respectively. Then, $s$ is further passed through the style classifier $f_c$. To enable $E_s$ and $C_s$ to jointly learn a style-aware space, we employ triplet loss~\cite{dong2018triplet} and the cross-entropy loss in addition to the loss functions used in~\citet{van2017neural} during training:
	\begin{equation}
		\begin{aligned}
			\label{eq:1}
			L_s &=\left\|\operatorname{sg}[f_s]-q_s\right\|
			+\left\|\operatorname{sg}\left[q_s\right]-f_s\right\|\\
			&+\alpha_\text{trip }\text{max}\{\left\|s - s^p\right\|_2-\left\|s - s^n\right\|_2 + \gamma,0\}\\
			&+\alpha_\text{c}\sum_{c=1}^M y_c log(f_c(s)) + \|\hat{\beta^s_{1:T'}}-D_s(q_s)\|,
		\end{aligned}
	\end{equation}
	where $s^p$ and $s^n$ donate the style code extracted from the videos with the same and different style with input style clip, $M$ is the number of the styles, $y_c$ refers to the ground truth style class of $\beta^s$ and $\operatorname{sg}[\cdot]$ stands for a stop-gradient operation. Once the module is well-trained, the style codebook serves as a prior to guarantee high-fidelity style speaking extracting when processing an arbitrary style clip. The pipeline of style extraction is displayed in left Figure~\ref{fig:overview} (b).

	\subsection{Stylized Facial Motion Synthesis with HyperStyle}
	% \label{subsec:hyperstyle}
	We design an audio encoder $E_a$ to extract $z^a_{1:T}$ from the audio clip $a_{1:T}$ and source parameter $\beta^r$, which is extracted from one frame of corresponding video $V^r$. In order to ensure the synchronization of the mouth and audio in the resulting video while preserving the consistent speaking style with the style clip, we design a residual motion stylized module. The module comprises a canonical branch $\phi_c$ for enabling $z^a_{1:T}$ to integrate audio-driven lip motion information with canonical style, and style-specific branch $\phi_s$ for predicting the offset from the canonical one to stylized one. Both branches accept $z^a_{1:T}$ as input, while the style-specific branch takes an additional style code, to specify the desired speaking style. 
	% Nevertheless, due to the significant disparity among various speaking styles, relying on a uniform network weight becomes inadequate for effectively managing the diverse range~\cite{tao2022style, mor2018universal, mason2018few}. 
	Subsequently, we explore a HyperStyle $H$ to efficiently produce the adapted style-attention weights for the branch. In particular, the style-specific branch is based on a 6-layer LSTM and our HyperStyle $H$ predicts a set of offsets with respect to the original weights of the middle 4 layers. To assist the $H$ in inferring the desired modifications, we pass style code $s$ as the input. The process can be expressed as:
	\begin{equation}
		z^s_{1:T} = \phi_c(z^a_{1:T}) + \phi_s(z^a_{1:T}, s, H(s)),
	\end{equation}
	where the latent code $z^s_{1:T}$ and style code $s$ are further fed into the decoder $D$ to generate stylized 3D coefficients $\hat{\beta^s_{1:T}}$.
	
	\begin{figure}[t]
		\centering
		\includegraphics[width=0.9\linewidth]{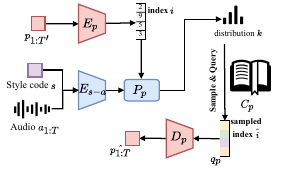}
		\caption{The pipeline of pose generator $G_p$.}
		\label{fig:head_pose}
	\end{figure}
	
	We supervise our model in terms of lip-sync and stylized expression, respectively. As for the former, we follow the self-supervised training strategy, in which the coefficients are predicted from a style clip, corresponding audio and a reference image selected from other video frames of the same speaker. To avoid the possible misalignment problem between audio and ground truth coefficients, we adopt the Soft-DTW loss $L_\text{rec}$~\cite{chen2023improving} to minimize the difference between the predicted coefficients $\hat{\beta_{1:T}}$ and ground truth $\beta_{1:T}$. On the other hand, to guide the framework to generate vivid speaking styles, we employ the triplet loss $L_\text{trip}$ and the pre-trained style classifier $f_c$ to constrain the generated coefficients (donated as $L_\text{style1}$) in Equation~\ref{eq:1}. Sparked by the success of style transfer~\cite{tao2022style}, a style discriminator is presented to further enhance the temporal speaking style. The style discriminator receives the coefficient sequence and ground truth style class as input and determines whether the coefficients perform with correct style or not from the aspects of feature and temporal (donated as $L_\text{style2}$). To sum up, the total loss to train our SAAS is calculated by the weighted sum of the above loss functions:
	\begin{equation}
		\label{eq:4}
		L_\text{total} = L_\text{rec} + \alpha_\text{trip} L_\text{trip}
		+ \alpha_\text{style1} L_\text{style1}+\alpha_\text{style2} L_\text{style2}, 
	\end{equation}
	where we move the detailed derivation of the formulas into the supplementary material due to the limited space.
	
	\begin{table*}[t]
		\centering
		\resizebox{\linewidth}{!}{
			\begin{tabular}{c|ccccc|ccccc}
				\toprule
				\multicolumn{1}{c}{\multirow{2}[4]{*}{\textbf{Method}}} & \multicolumn{5}{c}{\textbf{MEAD}~\cite{wang2020mead}} & \multicolumn{5}{c}{\textbf{HDTF}~\cite{zhang2021flow}}\\
				\cmidrule(lr){2-6}  \cmidrule(lr){7-11}  \multicolumn{1}{c}{} & \multicolumn{1}{c}{SSIM$\uparrow$} & \multicolumn{1}{c}{FID$\downarrow$} & \multicolumn{1}{c}{M-LMD$\downarrow$} & \multicolumn{1}{c}{F-LMD$\downarrow$} & \multicolumn{1}{c}{$\text{Sync}_\text{conf}\uparrow$} & \multicolumn{1}{c}{SSIM$\uparrow$} & \multicolumn{1}{c}{FID$\downarrow$} & \multicolumn{1}{c}{M-LMD$\downarrow$} & \multicolumn{1}{c}{F-LMD$\downarrow$} & \multicolumn{1}{c}{$\text{Sync}_\text{conf}\uparrow$}
				\\
				
				\midrule
				\multirow{1}[2]{*}{MakeItTalk} & 0.618 &73.064     &4.314     &   4.778   &  1.719    &  0.648   & 23.124    &   5.061   & 5.193     & 1.888   \\
				\multirow{1}[2]{*}{Wav2Lip} & 0.635 & 86.812   &4.146     &   4.271  &  3.275    &   0.729  &19.349    &   4.571  &4.667      & \textbf{5.077}   \\
				\multirow{1}[2]{*}{Audio2Head} & 0.609 & 84.315     &5.636     &  5.997   &  2.796   &   0.610   & 31.503    &   6.467    &6.263      &1.930   \\
				\multirow{1}[2]{*}{PC-AVS} & 0.588 & 95.913    &6.592     &   6.969   &  2.837    &   0.431   & 128.806   &   7.827   &7.675    &3.661   \\
				
				\multirow{1}[2]{*}{AVCT} & 0.656 & 83.574    &4.883     &  4.676   &  2.946    &   0.633   &44.636    &   5.393  & 5.376      & 4.231 \\
				\multirow{1}[2]{*}{SadTalker} & 0.636& 88.750    &3.637      &   3.736  &  3.135    & 0.697   & \textbf{18.317}    &   3.039   &3.143     & 3.157  \\
				
				\midrule
				
				\multirow{1}[2]{*}{EAMM} & 0.624 & 83.396     &4.964    &   4.458   & 2.708    &  0.630   & 57.145    &  5.353  &5.946    &1.555  \\
				
				\multirow{1}[2]{*}{StyleTalk} &0.669 & 68.399    &3.361    &  3.262  &3.288   &  0.723  & 19.327   &  2.758  &2.448     & 2.445  \\
				\textbf{SAAS} & \textbf{0.683} & \textbf{59.718}     &\textbf{3.104}     &   \textbf{2.914}   &  \textbf{3.346 }   &   \textbf{0.732}   & 18.919    &   \textbf{2.588}   &\textbf{2.185}     & 2.717   \\ 
				\textbf{SAAS-V} & \textbf{0.830} & \textbf{40.862}     &\textbf{1.552}     &   \textbf{1.413}   & \textbf{3.490 }   &   \textbf{0.873}   & \textbf{9.057}    &   \textbf{1.412}   & \textbf{1.323}    & 3.645   \\ 
				\midrule
				\multirow{1}[2]{*}{GT} & 1.000 & 0.000     &0.000   &   0.000  &  3.590    &   1.000  & 0.000     &0.000   &   0.000     & 2.903  \\
				\bottomrule
				
			\end{tabular}%
		}
		\caption{Quantitative comparisons with state-of-the-art methods. We test each method on MEAD and HDTF datasets, and the best scores in each metric are highlighted in bold. The signages $"\uparrow"$ and  $"\downarrow"$ indicate higher and lower metric values for better results, respectively.}
		\label{tab:quantitative}%
	\end{table*}%

	\subsection{Head Pose Synthesis}
	% \label{subsec:headpose}
	Our pose generator $G_p$ involves inputs of driving audio and style code extracted from the style clip as shown in Figure~\ref{fig:head_pose}. During training, we learn a pose codebook $C_p$ to store the pose prior jointly with a pose encoder $E_p$ and decoder $D_p$ in a self-reconstruction manner. Inspired by \citet{dosovitskiy2020image}, a cross-modal Encoder $E_{s-a}$ is employed to fuse the information across $a_{1:T}$ and $s$. $p_{1:T'}$ is represented as a sequence of corresponding codebook indices $i$ following~\citet{ng2022learning}, and then passed through pose predictor $P_p$ along with the fused information. The output of $P_p$ is a distribution $k$ of pose codebook indices, ensuring diverse head poses that align with the audio rhythm and style code. Subsequently, we sample the index $\hat{i}$ of the codebook $C_p$ from the distribution $k$ and then retrieve the corresponding quantized element ${q_p}_i$. Incorporating the decoder $D_p$, the corresponding pose coefficients $\hat{p_{1:T}}$ are obtained.
	
	Along with the predicted expression $\hat{\beta^s_{1:T}}$ and reference image $I^r$, the realistic stylized videos with head motions are generated via a Face Render $R$~\cite{ren2021pirenderer}. Since stylized expressions are mainly expressed through eye and mouth regions~\cite{GaryFaigin1990TheAC}, we introduce three facial discriminators to enhance the style performance in the significant facial regions during training the Face Render. More details can be found in the supplementary materials.

	\subsection{Extension on Video-driven Style Transfer}
	We extend our SAAS to the domain of the facial motion style transfer, which edits the speaking style of the given video while maintaining lip synchronized with the original video. In particular, we replace the audio encoder $E_a$ with a video encoder $E_v$ and keep the rest of the network structure consistent with the audio-driven setting. $E_v$ maps the 3D coefficients $\beta^r_{1:T}$ extracted from the source video $V^r$ to $z^v_{1:T}$, which contain both the source mouth shape and source speaking style. In this way, the canonical branch is able to convert $z^v_{1:T}$ into the canonical and further transfer to $z^s_{1:T}$ with the target speaking style $s$.
	
	Since there are no two videos synchronized with the same audio in different styles, we cannot constrain our network with reconstruction loss. Instead, we introduce a cycle-reconstruction loss, where the source video is first transferred to target style $s_t$, and then make the transferred video $\hat{\beta_{1:T}}$ perform as the source style $s_r$ again:
	\begin{gather}
		\hat{\beta_{1:T}} = D(\phi_c(E_v(\beta^v_{1:T})) + \phi_s((E_v(\beta^v_{1:T}), s_t, H(s_t))))\notag \\
		\overline{\beta^v_{1:T}} = D(\phi_c(E_v(\hat{\beta^v_{1:T}})) + \phi_s((E_v(\hat{\beta^v_{1:T}}), s_r, H(s_r))))\notag \\
		L_\text{cyc} = \| \beta^v_{1:T} -  \overline  {\beta^v_{1:T}} \|_2,
	\end{gather}
	where $\overline{\beta^v_{1:T}}$ refers to cycle-reconstructed result. Based on the assumption that the relative distance from the upper lip to the lower lip of the source video and generated stylized video should be consistent to ensure the synchronization with the audio~\cite{sun2022continuously}, we employ the mouth loss $L_\text{mouth}$:
	
	\begin{equation}
		\begin{aligned}
			L_\text{mouth} = &\|(\text{Upp}(\text{Ver}(\beta^v_{1:T}))) - \text{Low}(\text{Ver}(\beta^v_{1:T})))- \\
			& (\text{Upp}(\text{Ver}(\hat{\beta^v_{1:T}})) - \text{Low}(\text{Ver}(\hat{\beta^v_{1:T}})))\|_2,
		\end{aligned}
	\end{equation}
	where $\text{Ver}(\cdot)$ donates the 3D mesh vertices reconstructed from 3DMM coefficients, $\text{Upp}(\cdot)$ and $\text{Low}(\cdot)$ refer to the index of the upper and low lip vertices. To summarize, the objective functions $L^v_\text{total}$ for video-driven motion transfer are:
	\begin{equation}
		\begin{aligned}
			L^v_\text{total} = &L_\text{mouth}+ \alpha_\text{cyc}L_\text{cyc} + 
			\alpha_\text{trip} L_\text{trip} \\
			&+ \alpha_\text{style1} L_\text{style1}+ 
			\alpha_\text{style2} L_\text{style2}  
		\end{aligned}
	\end{equation}

	\begin{figure*}[t]
		\centering
		\includegraphics[width=0.92\linewidth]{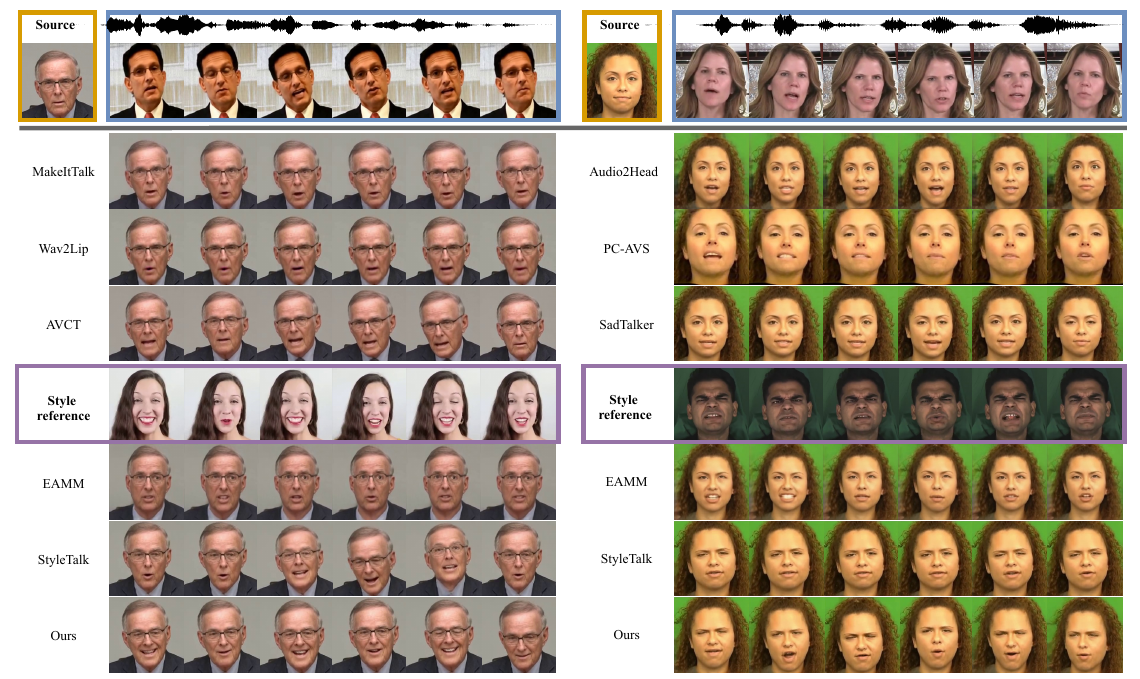}
		\caption{Qualitative comparisons with state-of-the-art methods. Top row shows the identity, driving audio and corresponding mouth ground truth. The purple row shows the style source clips.}
		\label{fig:compare}
	\end{figure*}

	\section{Experiments}
	\subsection{Experimental Settings}
	\paragraph{Datasets and Implementation Details.}
	Two public datasets are leveraged to train and test our proposed SAAS: MEAD~\cite{wang2020mead} and HDTF~\cite{zhang2021flow}. MEAD captures various emotional videos performed by 60 actors with 8 emotions and 3 levels, where MEAD is rich in expressions but only 159 sentences are recorded. HDTF consists of over 10k different sentences and 300 speakers collected from youtube website, which compensates for the limited MEAD subject and corpus. We implement our SAAS model with Pytorch. We set $w = 8$, $T' = 32$, $N = 500$ and $d_s = 256$. Model training and testing are conducted on 4 NVIDIA GeForce GTX 3090 with 24GB memory. Incorporating the Adaptive moment estimation (Adam) optimizer~\cite{kingma2014adam}, the style codebook $C_s$ and Style Encoder $E_s$ are pre-trained for 24 hours. Then, we froze weights of $C_s$ and $E_s$, and jointly train the whole network with the learning rate of 2e-4 for 500 and 300 epochs in audio-driven and video-driven settings, respectively.

	\paragraph{Comparison Setting.}
	We compare our SAAS against state-of-the-art (SOTA) methods including neutral talking face generation methods: MakeItTalk~\cite{zhou2020makelttalk}, Wav2Lip~\cite{prajwal2020lip}, Audio2Head~\cite{wang2021audio2head}, PC-AVS~\cite{zhou2021pose}, AVCT~\cite{wang2022one}, SadTalker~\cite{zhang2022sadtalker}, and stylized talking face generation methods: EAMM~\cite{ji2022eamm}, StyleTalk~\cite{ma2023styletalk}. The former focuses on the lip-synchronization with the same expression as the source image, while the latter additionally takes stylized expression into consideration by involving another style clip as input. We evaluate the generated videos using following metrics: SSIM~\cite{ZhouWang2004ImageQA}, FID~\cite{Seitzer2020FID}, M-LMD~\cite{chen2019hierarchical}, SyncNet~\cite{chung2017out} and F-LMD. SSIM and FID measure the distance between the generated video and ground truth from the image-level and feature level, respectively. Landmarks distances on the mouth (M-LMD) and the confidence score of SyncNet assess the synchronization between the generated lip motion and the input audio. Besides, F-LMD focuses on evaluating the similarity of the facial expression.
	% Due to space limitations, we move part of the experimental results to the supplementary material.
	
	\subsection{Experimental Results}
	\paragraph{Quantitative Results.}
	
	We quantitatively conduct the comparison experiments in a self-driven fashion. On HDTF, the first frame of each video and corresponding audio are used as the source image and driving audio for each method, and the videos are additionally fed into stylized talking face generation methods as speaking style reference. When testing on MEAD, the style clip and driving audio are selected similarly to HDTF, while the source image is sampled from a neutral video of the same subject. As for video-driven style transfer (donated as SAAS-V), since there is no ground truth for the stylized video produced by SAAS-V as aforementioned, we employ SAAS-V to first convert the speaking style of each video into neutral with the neutral video of the same speaker, and then stylize the neutralized results to the original speaking style, which are used to calculated metrics. All testing data are unseen during training.
	
	Table~\ref{tab:quantitative} summarizes the results of the quantitative comparison between ours and the SOTAs. We achieve the best performance in terms of all metrics on MEAD, and most metrics on HDTF. We obtain the higher score of SSIM and comparable scores of FID with SadTalker on HDTF, which demonstrates the superiority of our method in the image quality of the results. Wav2Lip achieves the highest score of $\text{Sync}_\text{conf}$ that even surpasses that of GT. The reason is probably that the confidence score of SyncNet is an important constraint when training Wav2Lip with SyncNet discriminator. In contrast, our method is not only similar to the  $\text{Sync}_\text{conf}$ score of GT, but also leads to the lowest disparity between the output and GT with regard to M-LMD. This indicates the precise lip-synchronization of our SAAS. Besides, SAAS-V significantly exceeds SOTAs among all metrics on both datasets, which suggests the effectiveness of our framework in both audio-driven and video-driven setups.

	\noindent \textbf{Qualitative Results.} The qualitative comparison to state-of-the-art methods is also conducted. We select source images, driving audios and style reference clips unseen in training set as the inputs for each method. Qualitative results are depicted in Figure~\ref{fig:compare}, where our method synthesizes realistic stylized talking face videos with accurate lip-synchronization and diverse head poses. Specifically, MakeItTalk~\cite{zhou2020makelttalk} fails to generate precise mouth shapes. Despite improved lip-sync, Wav2Lip~\cite{prajwal2020lip} produces the blur lower faces. Besides, both Audio2Head~\cite{wang2021audio2head} and PC-AVS~\cite{zhou2021pose} suffer from inconsistent identity with the source image. Furthermore, the head poses performed by AVCT~\cite{wang2022one} seem jittery and less continuous, making the output less realistic. Though SadTalker~\cite{zhang2022sadtalker} exempts the jitter, it neglects stylized expressions for generating realistic animation. While EAMM~\cite{ji2022eamm} takes speaking style into account, it struggles to imitate the expression of style reference and preserve the source identity. By extracting the speaking style in a discrete manner, rather than in a regression manner as in StyleTalk~\cite{ma2023styletalk}, we generate more explicit speaking styles in results.

	\noindent \textbf{User Study.}
	We conduct a user study to compare generation results with SOTAs. We produce 16 videos via each method, and invite 20 participants (10 males, 10 females) to score each video on a scale of 1 (worst) to 5 (best) in terms of lip synchronization and naturalness. Besides, participants are required to classify the speaking style performed by the results of stylized expression generation methods. The average scores are reported in Table~\ref{tab:userstudy}, demonstrating that our method outperforms other methods. More quantitative, qualitative and user study results can be found in the supplementary material.

	\begin{table}
		\centering
		\resizebox{\linewidth}{!}{
			\begin{tabular}{@{}l|ccccc@{}}
				\toprule
				Score/Method & PC-AVS & AVCT &EAMM  &StyleTalk  & SAAS \\
				\midrule
				Lip-Sync. & 3.16& 3.21 & 1.91 &3.38 & \textbf{3.58}  \\
				Naturalness & 1.85& 2.67 &1.47 & 2.98 & \textbf{3.28}   \\
				Style Accuracy & 16.8\% & 15.4\% & 46.2\%&63.4\% & \textbf{71.3}\%\\
				\bottomrule
			\end{tabular}
		}
		\caption{User study results. The score ranges from 1 to 5, and error bars imply the standard deviations.}
		\label{tab:userstudy}
	\end{table}
	
	\noindent \textbf{Ablation Study.}
	We further conduct the ablation study to investigate the contributions of different components in our SAAS. To be specific, the experiment arrangements can be concluded as: (1) w/o $C_s$: remove the style codebook and extract the style code in a regressive way. (2) $N=250$/ $N=750$: set the size of $C_s$ as 250 and 750. (3) w/o $H$: remove $H$ and use a unified style branch. (4) w/o $L_\text{trip}$/ w/o $L_\text{style1}$/ w/o $L_\text{style2}$: remove different loss functions. 
	
\begin{figure}[t]
		\centering
		\includegraphics[width=0.95\linewidth]{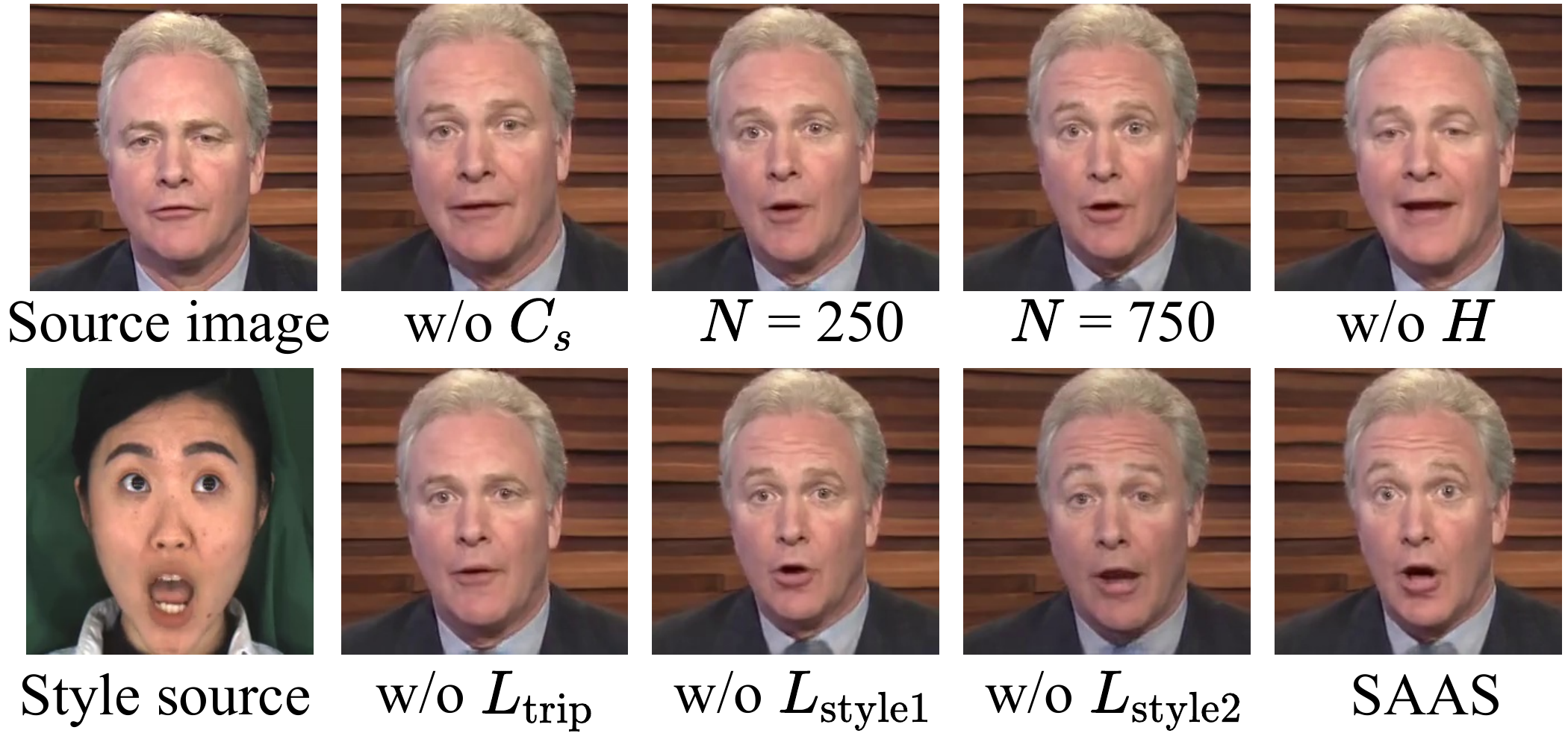}
		\caption{Visualization results of ablation study.}
		\label{fig:ablation}
	\end{figure}

	The visual results are presented in Figure~\ref{fig:ablation}. The speaking style of experiment (1) tends to be averaged without distinctive style, which is significantly improved with the assistance of $C_s$. This demonstrates that the discrete prior stored in the codebook facilitates the extraction of the speaking style. However, $N=250$ cannot expand enough space to include a wide variety of styles, while too huge space ($N=750$) makes the query process confusing. Therefore, we choose $N=500$ to achieve the best trade-off. Without the style-specific weights produced by $H$, experiment (3) performs poorly in lip-synchronization and speaking style. $L_\text{trip}$ in Equation~\ref{eq:1} mainly contributes to the speaking style extraction while $L_\text{trip}$ in Equation~\ref{eq:4} constrains the speaking style of the resulting video to be close to the positive sample and distant from the negative sample. $L_\text{style1}$ and $L_\text{style2}$ further enhance the low disparity between the styles in results and style source. Consequently, the contribution of each module is verified. The numerical results reported in Table~\ref{tab:ablation} also confirm our assumptions. Please refer to our supplementary materials for more ablation study results about different style codebook sizes and facial discriminators.
	
	\begin{table}
		\centering
		\resizebox{\linewidth}{!}{
			\begin{tabular}{@{}l|cccc@{}}
				\toprule
				Method/Score & SSIM $\uparrow$ &FID $\downarrow$ & M-LMD $\downarrow$ & F-LMD $\downarrow$\\
				\midrule
				w/o $C_s$ & 0.641 & 84.149 &3.714 & 3.545 \\
				$N=250$ & 0.664 &70.151 &  3.434 & 3.244  \\
				$N=750$ & 0.676 & 64.481 & 3.246 & 3.163\\
				w/o $H$ & 0.645 & 81.415 & 3.451 & 3.515\\
				w/o $L_\text{trip}$ & 0.659 & 75.045 & 3.283 & 3.421\\
				w/o $L_\text{style1}$ & 0.674 & 63.746 & 3.154 & 3.071\\
				w/o $L_\text{style2}$ & 0.676 & 67.418 & 3.215 & 3.062\\
				\midrule
				\textbf{Ours} & $\textbf{0.683}$ & $\textbf{59.718}$& $\textbf{3.104}$ & $\textbf{2.914}$\\
				\bottomrule
			\end{tabular}
		}
		\caption{Results for ablation study on MEAD dataset.}
		\label{tab:ablation}
	\end{table}

	\section{Conclusion}
	In this work, we present a novel \textbf{S}ay \textbf{A}nything with \textbf{A}ny \textbf{S}tyle (SAAS) to achieve stylized talking face generation. In contrast to the previous regressive way, we resort to discrete representation learning and construct a style codebook in a multi-task manner. Incorporating
	it, an arbitrary style can be condensed as the combination of the most pertinent elements in prior. A HyperStyle is developed to modulate the weight of the style-specific branch to enable stylized motion conditioned on the extracted speaking style and driving audio. To perform stylized head motion, we design another pose codebook and a pose generator, where the former expands a finite proxy space for the quantized pose representation, while the latter fuses the style and audio to sample diverse head poses. Furthermore, we extend this framework to tackle the video-driven style editing task. Extensive experiments demonstrate the superiority of our method.

	\section{Acknowledgments}
	This work was supported by National Natural Science Foundation of China (NSFC, NO. 62102255) and Shanghai Municipal Science, Technology Major Project (No. 2021SHZDZX0102), the 2022 Hangzhou Key Science and Technology Innovation Program (No. 2022AIZD0054), and the Key Research and Development Program of Zhejiang Province (No. 2022C01011). We would like to thank Xinya Ji, Yifeng Ma and Zhiyao Sun for their generous help.
	
	\bibliography{SAAS}

\begin{thebibliography}{56}
\providecommand{\natexlab}[1]{#1}

\bibitem[{Alghamdi et~al.(2022)Alghamdi, Wang, Bulpitt, and Hogg}]{alghamdi2022talking}
Alghamdi, M.~M.; Wang, H.; Bulpitt, A.~J.; and Hogg, D.~C. 2022.
\newblock Talking Head from Speech Audio using a Pre-trained Image Generator.
\newblock In \emph{Proceedings of the 30th ACM International Conference on Multimedia}, 5228--5236.

\bibitem[{Blanz and Vetter(1999)}]{blanz1999morphable}
Blanz, V.; and Vetter, T. 1999.
\newblock A morphable model for the synthesis of 3D faces.
\newblock In \emph{Proceedings of the 26th annual conference on Computer graphics and interactive techniques}, 187--194.

\bibitem[{Chen et~al.(2020{\natexlab{a}})Chen, Cui, Liu, Li, Kou, Xu, and Xu}]{chen2020talking}
Chen, L.; Cui, G.; Liu, C.; Li, Z.; Kou, Z.; Xu, Y.; and Xu, C. 2020{\natexlab{a}}.
\newblock Talking-head generation with rhythmic head motion.
\newblock In \emph{European Conference on Computer Vision}, 35--51. Springer.

\bibitem[{Chen et~al.(2019)Chen, Maddox, Duan, and Xu}]{chen2019hierarchical}
Chen, L.; Maddox, R.~K.; Duan, Z.; and Xu, C. 2019.
\newblock Hierarchical cross-modal talking face generation with dynamic pixel-wise loss.
\newblock In \emph{Proceedings of the IEEE/CVF conference on computer vision and pattern recognition}, 7832--7841.

\bibitem[{Chen et~al.(2023)Chen, Ma, Liu, Tan, Lu, Yu, and Chen}]{chen2023improving}
Chen, Q.; Ma, Z.; Liu, T.; Tan, X.; Lu, Q.; Yu, K.; and Chen, X. 2023.
\newblock Improving Few-Shot Learning for Talking Face System with TTS Data Augmentation.
\newblock In \emph{ICASSP 2023-2023 IEEE International Conference on Acoustics, Speech and Signal Processing (ICASSP)}, 1--5. IEEE.

\bibitem[{Chen et~al.(2020{\natexlab{b}})Chen, Dai, Liu, Chen, Yuan, and Liu}]{chen2020dynamic}
Chen, Y.; Dai, X.; Liu, M.; Chen, D.; Yuan, L.; and Liu, Z. 2020{\natexlab{b}}.
\newblock Dynamic convolution: Attention over convolution kernels.
\newblock In \emph{Proceedings of the IEEE/CVF conference on computer vision and pattern recognition}, 11030--11039.

\bibitem[{Chung and Zisserman(2017)}]{chung2017out}
Chung, J.~S.; and Zisserman, A. 2017.
\newblock Out of time: automated lip sync in the wild.
\newblock In \emph{Computer Vision--ACCV 2016 Workshops: ACCV 2016 International Workshops, Taipei, Taiwan, November 20-24, 2016, Revised Selected Papers, Part II 13}, 251--263. Springer.

\bibitem[{Deng et~al.(2019)Deng, Yang, Xu, Chen, Jia, and Tong}]{deng2019accurate}
Deng, Y.; Yang, J.; Xu, S.; Chen, D.; Jia, Y.; and Tong, X. 2019.
\newblock Accurate 3d face reconstruction with weakly-supervised learning: From single image to image set.
\newblock In \emph{Proceedings of the IEEE/CVF conference on computer vision and pattern recognition workshops}, 0--0.

\bibitem[{Dieleman, van~den Oord, and Simonyan(2018)}]{dieleman2018challenge}
Dieleman, S.; van~den Oord, A.; and Simonyan, K. 2018.
\newblock The challenge of realistic music generation: modelling raw audio at scale.
\newblock \emph{Advances in Neural Information Processing Systems}, 31.

\bibitem[{Dong and Shen(2018)}]{dong2018triplet}
Dong, X.; and Shen, J. 2018.
\newblock Triplet loss in siamese network for object tracking.
\newblock In \emph{Proceedings of the European conference on computer vision (ECCV)}, 459--474.

\bibitem[{Dosovitskiy et~al.(2020)Dosovitskiy, Beyer, Kolesnikov, Weissenborn, Zhai, Unterthiner, Dehghani, Minderer, Heigold, Gelly et~al.}]{dosovitskiy2020image}
Dosovitskiy, A.; Beyer, L.; Kolesnikov, A.; Weissenborn, D.; Zhai, X.; Unterthiner, T.; Dehghani, M.; Minderer, M.; Heigold, G.; Gelly, S.; et~al. 2020.
\newblock An image is worth 16x16 words: Transformers for image recognition at scale.
\newblock \emph{arXiv preprint arXiv:2010.11929}.

\bibitem[{Ekman and Rosenberg(2005)}]{PaulEkman2005WhatTF}
Ekman, P.; and Rosenberg, E.~L. 2005.
\newblock What the face reveals : basic and applied studies of spontaneous expression using the facial action coding system (FACS).

\bibitem[{Faigin(1990)}]{GaryFaigin1990TheAC}
Faigin, G. 1990.
\newblock The Artist's Complete Guide to Facial Expression.

\bibitem[{Ha, Dai, and Le(2016)}]{ha2016hypernetworks}
Ha, D.; Dai, A.; and Le, Q.~V. 2016.
\newblock Hypernetworks.
\newblock \emph{arXiv preprint arXiv:1609.09106}.

\bibitem[{Ji et~al.(2022)Ji, Zhou, Wang, Wu, Wu, Xu, and Cao}]{ji2022eamm}
Ji, X.; Zhou, H.; Wang, K.; Wu, Q.; Wu, W.; Xu, F.; and Cao, X. 2022.
\newblock Eamm: One-shot emotional talking face via audio-based emotion-aware motion model.
\newblock In \emph{ACM SIGGRAPH 2022 Conference Proceedings}, 1--10.

\bibitem[{Ji et~al.(2021)Ji, Zhou, Wang, Wu, Loy, Cao, and Xu}]{ji2021audio}
Ji, X.; Zhou, H.; Wang, K.; Wu, W.; Loy, C.~C.; Cao, X.; and Xu, F. 2021.
\newblock Audio-driven emotional video portraits.
\newblock In \emph{Proceedings of the IEEE/CVF conference on computer vision and pattern recognition}, 14080--14089.

\bibitem[{Jo and Kim(2021)}]{jo2021practical}
Jo, Y.; and Kim, S.~J. 2021.
\newblock Practical single-image super-resolution using look-up table.
\newblock In \emph{Proceedings of the IEEE/CVF Conference on Computer Vision and Pattern Recognition}, 691--700.

\bibitem[{Kingma and Ba(2014)}]{kingma2014adam}
Kingma, D.~P.; and Ba, J. 2014.
\newblock Adam: A method for stochastic optimization.
\newblock \emph{arXiv preprint arXiv:1412.6980}.

\bibitem[{Li et~al.(2021)Li, Wang, Zhang, Ding, Zheng, Yu, and Fan}]{li2021write}
Li, L.; Wang, S.; Zhang, Z.; Ding, Y.; Zheng, Y.; Yu, X.; and Fan, C. 2021.
\newblock Write-a-speaker: Text-based emotional and rhythmic talking-head generation.
\newblock In \emph{Proceedings of the AAAI Conference on Artificial Intelligence}, volume~35, 1911--1920.

\bibitem[{Liang et~al.(2022)Liang, Pan, Guo, Zhou, Hong, Han, Han, Liu, Ding, and Wang}]{liang2022expressive}
Liang, B.; Pan, Y.; Guo, Z.; Zhou, H.; Hong, Z.; Han, X.; Han, J.; Liu, J.; Ding, E.; and Wang, J. 2022.
\newblock Expressive talking head generation with granular audio-visual control.
\newblock In \emph{Proceedings of the IEEE/CVF Conference on Computer Vision and Pattern Recognition}, 3387--3396.

\bibitem[{Ma et~al.(2023)Ma, Wang, Hu, Fan, Lv, Ding, Deng, and Yu}]{ma2023styletalk}
Ma, Y.; Wang, S.; Hu, Z.; Fan, C.; Lv, T.; Ding, Y.; Deng, Z.; and Yu, X. 2023.
\newblock StyleTalk: One-shot Talking Head Generation with Controllable Speaking Styles.
\newblock \emph{arXiv preprint arXiv:2301.01081}.

\bibitem[{Mason et~al.(2018)Mason, Starke, Zhang, Bilen, and Komura}]{mason2018few}
Mason, I.; Starke, S.; Zhang, H.; Bilen, H.; and Komura, T. 2018.
\newblock Few-shot learning of homogeneous human locomotion styles.
\newblock In \emph{Computer Graphics Forum}, volume~37, 143--153. Wiley Online Library.

\bibitem[{Ng et~al.(2022)Ng, Joo, Hu, Li, Darrell, Kanazawa, and Ginosar}]{ng2022learning}
Ng, E.; Joo, H.; Hu, L.; Li, H.; Darrell, T.; Kanazawa, A.; and Ginosar, S. 2022.
\newblock Learning to listen: Modeling non-deterministic dyadic facial motion.
\newblock In \emph{Proceedings of the IEEE/CVF Conference on Computer Vision and Pattern Recognition}, 20395--20405.

\bibitem[{Pan et~al.(2024)Pan, Tan, Cheng, Lin, Zeng, and Mitchell}]{pan2024expressive}
Pan, Y.; Tan, S.; Cheng, S.; Lin, Q.; Zeng, Z.; and Mitchell, K. 2024.
\newblock Expressive Talking Avatars.
\newblock \emph{IEEE Transactions on Visualization and Computer Graphics}.

\bibitem[{Pan et~al.(2023)Pan, Zhang, Cheng, Tan, Ding, Mitchell, and Yang}]{pan2023emotional}
Pan, Y.; Zhang, R.; Cheng, S.; Tan, S.; Ding, Y.; Mitchell, K.; and Yang, X. 2023.
\newblock Emotional Voice Puppetry.
\newblock \emph{IEEE Transactions on Visualization and Computer Graphics}, 29(5): 2527--2535.

\bibitem[{Park et~al.(2022)Park, Kim, Hong, Choi, and Ro}]{park2022synctalkface}
Park, S.~J.; Kim, M.; Hong, J.; Choi, J.; and Ro, Y.~M. 2022.
\newblock Synctalkface: Talking face generation with precise lip-syncing via audio-lip memory.
\newblock In \emph{Proceedings of the AAAI Conference on Artificial Intelligence}, volume~36, 2062--2070.

\bibitem[{Pataranutaporn et~al.(2021)Pataranutaporn, Danry, Leong, Punpongsanon, Novy, Maes, and Sra}]{pataranutaporn2021ai}
Pataranutaporn, P.; Danry, V.; Leong, J.; Punpongsanon, P.; Novy, D.; Maes, P.; and Sra, M. 2021.
\newblock AI-generated characters for supporting personalized learning and well-being.
\newblock \emph{Nature Machine Intelligence}, 3(12): 1013--1022.

\bibitem[{Prajwal et~al.(2020)Prajwal, Mukhopadhyay, Namboodiri, and Jawahar}]{prajwal2020lip}
Prajwal, K.; Mukhopadhyay, R.; Namboodiri, V.~P.; and Jawahar, C. 2020.
\newblock A lip sync expert is all you need for speech to lip generation in the wild.
\newblock In \emph{Proceedings of the 28th ACM International Conference on Multimedia}, 484--492.

\bibitem[{Rebuffi, Bilen, and Vedaldi(2017)}]{rebuffi2017learning}
Rebuffi, S.-A.; Bilen, H.; and Vedaldi, A. 2017.
\newblock Learning multiple visual domains with residual adapters.
\newblock \emph{Advances in neural information processing systems}, 30.

\bibitem[{Ren et~al.(2021)Ren, Li, Chen, Li, and Liu}]{ren2021pirenderer}
Ren, Y.; Li, G.; Chen, Y.; Li, T.~H.; and Liu, S. 2021.
\newblock Pirenderer: Controllable portrait image generation via semantic neural rendering.
\newblock In \emph{Proceedings of the IEEE/CVF International Conference on Computer Vision}, 13759--13768.

\bibitem[{Ronneberger, Fischer, and Brox(2015)}]{ronneberger2015u}
Ronneberger, O.; Fischer, P.; and Brox, T. 2015.
\newblock U-net: Convolutional networks for biomedical image segmentation.
\newblock In \emph{Medical Image Computing and Computer-Assisted Intervention--MICCAI 2015: 18th International Conference, Munich, Germany, October 5-9, 2015, Proceedings, Part III 18}, 234--241. Springer.

\bibitem[{Safari, India, and Hernando(2020)}]{safari2020self}
Safari, P.; India, M.; and Hernando, J. 2020.
\newblock Self-attention encoding and pooling for speaker recognition.
\newblock \emph{arXiv preprint arXiv:2008.01077}.

\bibitem[{Seitzer(2020)}]{Seitzer2020FID}
Seitzer, M. 2020.
\newblock {pytorch-fid: FID Score for PyTorch}.
\newblock \url{https://github.com/mseitzer/pytorch-fid}.
\newblock Version 0.3.0.

\bibitem[{Sinha et~al.(2021)Sinha, Biswas, Yadav, and Bhowmick}]{SanjanaSinha2022EmotionControllableGT}
Sinha, S.; Biswas, S.; Yadav, R.; and Bhowmick, B. 2021.
\newblock Emotion-Controllable Generalized Talking Face Generation.
\newblock In \emph{International Joint Conference on Artificial Intelligence}. IJCAI.

\bibitem[{Sun et~al.(2022)Sun, Wen, Lv, Sun, Zhang, Wang, and Liu}]{sun2022continuously}
Sun, Z.; Wen, Y.-H.; Lv, T.; Sun, Y.; Zhang, Z.; Wang, Y.; and Liu, Y.-J. 2022.
\newblock Continuously Controllable Facial Expression Editing in Talking Face Videos.
\newblock \emph{arXiv preprint arXiv:2209.08289}.

\bibitem[{Tan, Ji, and Pan(2023)}]{tan2023emmn}
Tan, S.; Ji, B.; and Pan, Y. 2023.
\newblock EMMN: Emotional Motion Memory Network for Audio-driven Emotional Talking Face Generation.
\newblock In \emph{Proceedings of the IEEE/CVF International Conference on Computer Vision}, 22146--22156.

\bibitem[{Tan et~al.(2021)Tan, Tang, Peng, Xiao, Zu, Wu, Zhou, and Wang}]{tan2021incorporating}
Tan, S.; Tang, P.; Peng, X.; Xiao, J.; Zu, C.; Wu, X.; Zhou, J.; and Wang, Y. 2021.
\newblock Incorporating isodose lines and gradient information via multi-task learning for dose prediction in radiotherapy.
\newblock In \emph{Medical Image Computing and Computer Assisted Intervention--MICCAI 2021: 24th International Conference, Strasbourg, France, September 27--October 1, 2021, Proceedings, Part VII 24}, 753--763. Springer.

\bibitem[{Tao et~al.(2022)Tao, Zhan, Chen, and van~de Panne}]{tao2022style}
Tao, T.; Zhan, X.; Chen, Z.; and van~de Panne, M. 2022.
\newblock Style-ERD: responsive and coherent online motion style transfer.
\newblock In \emph{Proceedings of the IEEE/CVF Conference on Computer Vision and Pattern Recognition}, 6593--6603.

\bibitem[{Tian, Yuan, and Liu(2019)}]{tian2019audio2face}
Tian, G.; Yuan, Y.; and Liu, Y. 2019.
\newblock Audio2face: Generating speech/face animation from single audio with attention-based bidirectional lstm networks.
\newblock In \emph{2019 IEEE international conference on Multimedia \& Expo Workshops (ICMEW)}, 366--371. IEEE.

\bibitem[{Van Den~Oord, Vinyals et~al.(2017)}]{van2017neural}
Van Den~Oord, A.; Vinyals, O.; et~al. 2017.
\newblock Neural discrete representation learning.
\newblock \emph{Advances in neural information processing systems}, 30.

\bibitem[{Vougioukas, Petridis, and Pantic(2020)}]{vougioukas2020realistic}
Vougioukas, K.; Petridis, S.; and Pantic, M. 2020.
\newblock Realistic speech-driven facial animation with gans.
\newblock \emph{International Journal of Computer Vision}, 128(5): 1398--1413.

\bibitem[{Wang et~al.(2020)Wang, Wu, Song, Yang, Wu, Qian, He, Qiao, and Loy}]{wang2020mead}
Wang, K.; Wu, Q.; Song, L.; Yang, Z.; Wu, W.; Qian, C.; He, R.; Qiao, Y.; and Loy, C.~C. 2020.
\newblock Mead: A large-scale audio-visual dataset for emotional talking-face generation.
\newblock In \emph{Computer Vision--ECCV 2020: 16th European Conference, Glasgow, UK, August 23--28, 2020, Proceedings, Part XXI}, 700--717. Springer.

\bibitem[{Wang et~al.(2021)Wang, Li, Ding, Fan, and Yu}]{wang2021audio2head}
Wang, S.; Li, L.; Ding, Y.; Fan, C.; and Yu, X. 2021.
\newblock Audio2Head: Audio-driven One-shot Talking-head Generation with Natural Head Motion.
\newblock In \emph{International Joint Conference on Artificial Intelligence}. IJCAI.

\bibitem[{Wang et~al.(2022)Wang, Li, Ding, and Yu}]{wang2022one}
Wang, S.; Li, L.; Ding, Y.; and Yu, X. 2022.
\newblock One-shot talking face generation from single-speaker audio-visual correlation learning.
\newblock In \emph{Proceedings of the AAAI Conference on Artificial Intelligence}, volume~36, 2531--2539.

\bibitem[{Wang et~al.(2004)Wang, Bovik, Sheikh, and Simoncelli}]{ZhouWang2004ImageQA}
Wang, Z.; Bovik, A.~C.; Sheikh, H.~R.; and Simoncelli, E.~P. 2004.
\newblock Image quality assessment: from error visibility to structural similarity.
\newblock \emph{IEEE Transactions on Image Processing}.

\bibitem[{Wen et~al.(2023)Wen, Xiao, Tan, Wu, Zhou, Peng, and Wang}]{wen2023transformer}
Wen, L.; Xiao, J.; Tan, S.; Wu, X.; Zhou, J.; Peng, X.; and Wang, Y. 2023.
\newblock A Transformer-Embedded Multi-Task Model for Dose Distribution Prediction.
\newblock \emph{International Journal of Neural Systems}, 2350043--2350043.

\bibitem[{Xing et~al.(2023)Xing, Xia, Zhang, Cun, Wang, and Wong}]{xing2023codetalker}
Xing, J.; Xia, M.; Zhang, Y.; Cun, X.; Wang, J.; and Wong, T.-T. 2023.
\newblock CodeTalker: Speech-Driven 3D Facial Animation with Discrete Motion Prior.
\newblock \emph{arXiv preprint arXiv:2301.02379}.

\bibitem[{Ye et~al.(2022)Ye, Xia, Yi, Zhang, Lai, Huang, Zhang, and Liu}]{ye2022audio}
Ye, Z.; Xia, M.; Yi, R.; Zhang, J.; Lai, Y.-K.; Huang, X.; Zhang, G.; and Liu, Y.-j. 2022.
\newblock Audio-driven talking face video generation with dynamic convolution kernels.
\newblock \emph{IEEE Transactions on Multimedia}.

\bibitem[{Zamora~Esquivel et~al.(2019)Zamora~Esquivel, Cruz~Vargas, Lopez~Meyer, and Tickoo}]{zamora2019adaptive}
Zamora~Esquivel, J.; Cruz~Vargas, A.; Lopez~Meyer, P.; and Tickoo, O. 2019.
\newblock Adaptive convolutional kernels.
\newblock In \emph{Proceedings of the IEEE/CVF International Conference on Computer Vision Workshops}, 0--0.

\bibitem[{Zhang et~al.(2021{\natexlab{a}})Zhang, Ni, Fan, Li, Zeng, Budagavi, and Guo}]{zhang20213d}
Zhang, C.; Ni, S.; Fan, Z.; Li, H.; Zeng, M.; Budagavi, M.; and Guo, X. 2021{\natexlab{a}}.
\newblock 3d talking face with personalized pose dynamics.
\newblock \emph{IEEE Transactions on Visualization and Computer Graphics}.

\bibitem[{Zhang et~al.(2022)Zhang, Cun, Wang, Zhang, Shen, Guo, Shan, and Wang}]{zhang2022sadtalker}
Zhang, W.; Cun, X.; Wang, X.; Zhang, Y.; Shen, X.; Guo, Y.; Shan, Y.; and Wang, F. 2022.
\newblock SadTalker: Learning Realistic 3D Motion Coefficients for Stylized Audio-Driven Single Image Talking Face Animation.
\newblock \emph{arXiv preprint arXiv:2211.12194}.

\bibitem[{Zhang et~al.(2023)Zhang, Hu, Deng, Fan, Lv, and Ding}]{zhang2023dinet}
Zhang, Z.; Hu, Z.; Deng, W.; Fan, C.; Lv, T.; and Ding, Y. 2023.
\newblock DINet: Deformation Inpainting Network for Realistic Face Visually Dubbing on High Resolution Video.

\bibitem[{Zhang et~al.(2021{\natexlab{b}})Zhang, Li, Ding, and Fan}]{zhang2021flow}
Zhang, Z.; Li, L.; Ding, Y.; and Fan, C. 2021{\natexlab{b}}.
\newblock Flow-guided one-shot talking face generation with a high-resolution audio-visual dataset.
\newblock In \emph{Proceedings of the IEEE/CVF Conference on Computer Vision and Pattern Recognition}, 3661--3670.

\bibitem[{Zhou et~al.(2019)Zhou, Liu, Liu, Luo, and Wang}]{zhou2019talking}
Zhou, H.; Liu, Y.; Liu, Z.; Luo, P.; and Wang, X. 2019.
\newblock Talking face generation by adversarially disentangled audio-visual representation.
\newblock In \emph{Proceedings of the AAAI conference on artificial intelligence}, volume~33, 9299--9306.

\bibitem[{Zhou et~al.(2021)Zhou, Sun, Wu, Loy, Wang, and Liu}]{zhou2021pose}
Zhou, H.; Sun, Y.; Wu, W.; Loy, C.~C.; Wang, X.; and Liu, Z. 2021.
\newblock Pose-controllable talking face generation by implicitly modularized audio-visual representation.
\newblock In \emph{Proceedings of the IEEE/CVF conference on computer vision and pattern recognition}, 4176--4186.

\bibitem[{Zhou et~al.(2020)Zhou, Han, Shechtman, Echevarria, Kalogerakis, and Li}]{zhou2020makelttalk}
Zhou, Y.; Han, X.; Shechtman, E.; Echevarria, J.; Kalogerakis, E.; and Li, D. 2020.
\newblock MakeIttalk: speaker-aware talking-head animation.
\newblock \emph{ACM Transactions On Graphics (TOG)}, 39(6): 1--15.

\end{thebibliography}
	
\end{document}